# Experimentally Comparing Uncertain Inference Systems to Probability

Ben P. Wise
Thayer School, Dartmouth
Hanover, NH 03755

## 1. Introduction

Uncertainty is a pervasive feature of the domains in which expert systems are supposed to function. There are several mechanisms for handling uncertainty, of which the oldest and most widely used is probability theory. It is also the only one which is derived from a formal description of rational behavior [Savage 54]. There are many axiomatic arguments that no other system can do better than probability theory in terms of *results* [de Finetti74]. For use in pattern-directed inference systems, or rule-based inference engines, artificial intelligence researchers have favored others (such as fuzzy set theory Dempster-Shafer theory, or the algorithms used in the Prospector or Mycin expert systems), largely for reasons of simplicity, speed, low data requirements, and explainability. We will present techniques with which to measure how well these alternatives approximate the results of probability theory, to assess how well they perform by those measures, and to find out what underlying features of a problem cause strong or weak performance.

Because the amount of data required to fully specify a probability distribution is enormous for problems of practical size, some technique must be used to estimate a distribution when only partial information is given. Moreover, there is no formally correct Bayesian way to directly handle the type of uncertain information which expert systems must use. We give intuitive and axiomatic arguments that fitting maximum entropy priors and using minimum cross entropy updating are the most appropriate, or the most nearly Bayesian, ways to meet both requirements. Hence, we will compare UIS's to minimum cross-entropy updating (MXE). We have concentrated on an analysis of the system used in MYCIN [Shortliffe 76]. Its operations have been analyzed to elucidate both which basic problem-features affect, or bias, the answers, and the directions of the biases. Series of experiments have been done on test cases to find out how these biases affect the performance of the uncertain inference systems. We present and discuss both the motivation and design of our analysis techniques, and the specific structures which were found to have strong effects on bias and on performance.

## 2. Outline of Our Method

The basic goal of our work is to suggest what conditions produce significant differences in the outputs of uncertain inference systems (UIS's). The emphasis is on comparing the outputs, not the simplicity, explicability, or ease of construction of the UIS itself. We will present the method by giving the rationale for each part of the experimental procedure, as given in figure 2-1. The bottom is where the ME/MXE inference is performed while the inference of the UIS being explored is done in the top row.

Conversions: To answer questions about differences in performance, a common interpretation of the inputs and results is required to make them commensurable. Presumably, the ultimate purpose of any expert system is to lead to better decisions - either directly by the system or indirectly by the human user. If two different representations of uncertainty lead to making the same decision, then they are operationally equivalent. According to Bayesian Decision theory, decisions reveal beliefs about the outcomes on which the decisions are based. Even if the decision-maker does not think in probabilistic terms, if he chooses coherently he will act as though he did. In principle, if a non-probabilistic approach to uncertainty provides a theory of how to make both inferences and decisions based on its representations of uncertainty, then this would imply an operational equivalence between beliefs expressed in the probabilistic and non-probabilistic forms, where they produced the same decision. This would allow direct comparison of the representations. Notably, however, the non-probabilistic approaches do not provide agreed upon decision strategies, and so this is unfortunately impossible.

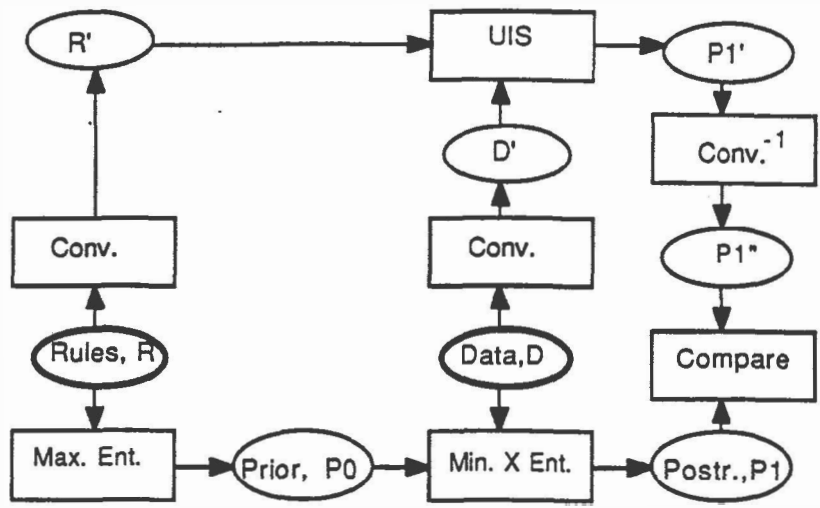

**Figure 2-1:** Basic Experiment Design for Comparisons

Nevertheless, there are obvious, simple ways to make transformations from to probability for at least some UIS's. For MYC, the conversions are taken from the definitions of Certainty Factors [Shortliffe 76]: $CF=1$ means true, so $p_1(x)=1$; $CF=0$ means no evidence, so we stay at the prior probability, $p_1(x)=p_0(x)$; and $CF=-1$ means definitely false, so $p_1(x)=0$. Values between these three points are found by piecewise linear interpolation. As these definitions were provided by the system's designers, we regard them as being quite accurate statements of the intended interpretation. One simplified version of probability always assumes that antecedents are conditionally independent, given their common conclusion; we will denote this version as CI. Using conversions, we can state inference rules and data in terms of probabilities, convert them to an alternative representation, and remain confident that the two still represent the same knowledge. At the end of an experiment, we invert the conversion to change the UIS's conclusions to probabilities, for direct comparison.

UIS Inference: The rule-strengths and data obtained by the above conversions are given to the UIS currently being tested, and propagated up a rule-tree according to the particular UIS's rules for *"and"*, *"or"*, *modus ponens* evaluating the reliability of evidence, or combining the results of different rules with the same consequent. The *"and"*, *"or"*, *"not"*, and implication (or *modus ponens*) operations for MYC are given in equation 1 The *"and"* and *"or"* operations are assumed; the *"not"* operation is easily derived.

$$cf(A_1 \& A_2) = \min[cf(A_1), cf(A_2)] \qquad cf(A_1 \text{ or } A_2) = \max[cf(A_1), cf(A_2)] \qquad (1)$$

$$cf(\neg A) = -cf(A) \qquad cf_1(C) = cf(C|A)cf_1(A), \text{ if } cf_1(A) \geq 0, \text{ 0 otherwise}$$

One should note that this last rule for *modus ponens* contradicts what Bayesian theory requires. That is, if A's being true increases $p_1(C)$ above its prior value, then A's being false must lower $p_1(C)$ below its prior value. But in MYC, A's being false causes C to stay at its prior probability. We define a UIS which is identical to MYC, except that it also repsonds to negative CF's for A. Because we call it Two-Sided Mycin, label it TSM.

The definition of a MYC rule-strength is quite similar to that for the CF of a proposition, except that we look at the change in the conclusion's probability, given the antecedant, relative to its prior probabilitiy. That is, we compare $p_0(C|A)$ and $p_0(C)$. The idea is that the CF's on the input evidence represent the change in belief attributable to some external evidence available only to the expert, while the rule CF's represent the change in belief in C which is attributable to changed belief in A, when C and A are both in the expert system.

When two rules refer to the same consequent, C, we use the rule in equation 2 for combining their results. The result of rule $R_1$ is denoted by $x$, and the result of rule $R_2$ by $y$, and the resulting CF in C by $z$. Surprisingly, this rather complicated rule is commutative and associative, so it can be applied to an arbitrary number of rules, in arbitrary order.

$$z = x + y - xy, \text{ if } x \geq 0, y \geq 0 \quad (2)$$
$$= x + y + xy, \text{ if } x \leq 0, y \leq 0$$
$$= (x + y) / [1 - \min(|x|, |y|)], \text{ if mixed signs.}$$

We will very briefly discuss another UIS, called CI, which is a simplified version of probability which represents all its rules as having their antecedants conditionally independent, given the consequent. Propositions $A_1$ and $A_2$ are conditionally independent, given C, if equation 3 holds.
$$p(A_1 \& A_2 | C) = p(A_1 | C) p(A_2 | C) \quad (3)$$

Intuitively, this states that the factors which affect $A_1$ and $A_2$ are unrelated, once we have compensated for all those which affect their relationship to C. Of course, it is not always possible for $A_1$ and $A_2$ to be related only through that one C, particularly if they themselves share a common antecedent, but the approximate formula can be used nonetheless. To do *modus ponens*, we use equation 4, which states that the updated odds-ratio for C is determined from its old odds-ratio, and the product of a series of factors, one for each antecedent. An approximate formula for interpolation is used when the $p(A_i)$ are between 0 and 1, as in equation 5.

$$\frac{p_1(C|A_1 \& A_2)}{p_1(\neg C|A_1 \& A_2)} = \frac{p_0(A_1|C)}{p_0(\neg A_1|C)} \frac{p_0(A_1|C)}{p_0(\neg A_1|C)} \frac{p_0(C)}{p_0(\neg C)} \quad (4)$$

$$\frac{p_1(C)}{p_1(\neg C)} = \frac{p_0(A_1|C)a + p_0(\neg A_1|C)(1-a)}{p_0(A_1|\neg C)a + p_0(\neg A_1|\neg C)(1-a)} \frac{p_0(A_2|C)b + p_0(\neg A_2|C)(1-b)}{p_0(A_2|\neg C)b + p_0(\neg A_2|\neg C)(1-b)} \frac{p_0(C)}{p_0(\neg C)} \quad (5)$$
where $p_1(A_1)=a$, and $p_1(A_2)=b$ and $p_1(C)=p_1(C|a \& b)$

ME/MXE Inference: The ME/MXE inference is a simple generalization of Bayesian conditioning. There are no special rules for *"and"*, *"or"*, *modus ponens*, evaluating the reliability of evidence, or combining the results of different rules with the same consequent. The first two "operations" are done by reading the appropriate numbers off the posterior. The last three "operations" happen automatically when we update. We must

321

define a particular rule for uncertain inference which will appear several times - Jeffrey's Rule. Where $x$ is some arbitrary event, and $y_i$ are an exhaustive and exclusive set of alternatives, equation 6 must hold for any distribution. In words, it says that our prior probability of any event $x$ must be a combination of our probabilities for $x$ given other events, $y_i$, weighted by our prior probabilities for those events. One may extend equation 6 from a consistency requirement to an updating rule by forming equation 7, which is the usual definition of Jeffrey's Rule.

$$p_0(x) = \sum_i p_0(x|y_i) p_0(y_i) \tag{6}$$

$$p_0(x) = \sum_i p_0(x|y_i) p_1(y_i) \tag{7}$$

Equation 7 has several notable characteristics. It always results in another coherent distribution. If $x$ is quite likely given a particular $y_i$ then raising the probability of that $y_i$ will raise the probability of $x$, so it exhibits an intuitively pleasing sensitivity. The new probabilities may be specified only over an exhaustive and exclusive set of events. The requirement of exhaustiveness is not really a problem - if new probabilities are given over a non-exhaustive set of events, one need only specify that the remaining probability fall over the remaining events, and so obtain an exhaustive set of events.

As discussed in [Diaconis 82], Jeffrey's Rule is a special case of many other generalized updating methods. In particular, it is a special case of the iterative proportional fitting procedure (IPFP), which is in turn a special case of minimum cross entropy (MXE) updating as proved in [Wise 86]. The IPFP is an extension to the case where the intersection of $y_i$ and $y_j$ is non-empty, simply by using Jeffrey's Rule for each $y_i$, over and over, until the result no longer changes significantly. The IPFP is the only form of MXE which we will need in this paper. However, we will briefly review the general justification of MXE.

Surprisingly, the maximum entropy estimation of priors and the minimum cross entropy updating method are completely defined by several weak properties, as proven by Shore and Johnson [Shore and Johnson 80]. They proved that any method for forming or updating a prior which possessed these properties would give the same answer as maximum entropy or minimum cross entropy, respectively. Thus, ME/MXE analysis has been singled out as not just one more heuristic among many, but as the only one which has these four simple consistency properties. One should note especially that it has been singled out whether it is considered as a "stand alone" method or as a computationally feasible approximation to Bayesian analysis. It is quite important that this proof does not specify that only the general MXE algorithms will satisfy these four properties. It states only that the *result* will be the same if we use any other method which also satisfies them.

The four properties are defined so as to make the answer invariant when we re-phrase the input data in ways which are by definition equivalent. First, the method should give a single, unique answer for the distribution. Second, the answer should be invariant under 1-to-1 coordinate transformations of the data or the prior distribution. Third, the answer should be invariant under different ways of specifying the probabilities of events which are independent. That is, we should be able to give marginal distributions over $y$ and over $x$, specifying that they are independent, or to give the resulting joint distribution, and still get the same answer. This basically means that the inference system must understand what the word "independent" means. Lastly, the answer should be invariant under different ways of specifying the probabilities of sets of events which are independent. Many more arguments for and against ME/MXE, as well as many comparisons to standard Bayesian analysis, can be found in [Wise 86].

Comparisons: Given the conclusions of MXE and the conclusions (converted back to probabilities) of a UIS, we can then do several comparisons. Some are useful in elucidating when biases will be apparent and how large they will be, others are useful in estimating how significant those biases may be to the user. We can measure one form of bias by seeing if the UIS has shifted the probabilities by the same additive amount, or by the same multiplicative factor. We will denote the prior probability by $p_0$, the UIS estimate by $p_{U,1}$, and the MXE estimate by $p_{M,1}$. We can compare the UIS-shift, defined by $\Delta_U = p_{U,1} - p_0$, to the MXE-shift, $\Delta_M = p_{M,1} - p_0$. We can run a whole series of tests, for different data, and do a regression of the UIS-shift against the MXE-shift, to see if there is consistent over- or under-response to the data, and if there is a constant bias in one direction, independent of the data.

We can also do comparisons of performance by evaluating how close the UIS's estimate is to the MXE answer. Because different $p_i$'s allow different margins for error, we compare the actual error to a perfect match, random guessing, or the worst possible match. This can be done either for the absolute error or the squared error; we will use squared error. For MYC and TSM, with random guessing of CF's over the interval [+1,-1], the average of the expected error in an estimate is given in equation 8. We may use the squared error to define a normalized performance measure, $\zeta$ which is 1 if there is zero error, 0 if the squared error is the expected value $\mu(\varepsilon^2)$, and -1 if the squared error is the largest possible, with linear interpolation between.

$$\mu(\varepsilon^2 | p_{M,1}, p_0) = \frac{2p_0^2 - 6p_{M,1}p_0 + 6p_{M,1}^2 + 1 + p_0 - 3p_{M,1}}{6} \tag{8}$$

### 3. Outline of Biases in MYC and TSM

In this section, our main concern will be to assess the biases which Mycin (MYC) displays. We will also briefly discuss the modified version, TSM. As discussed earlier, the standard of comparison will be ME/MXE analysis. Our general conclusions will be that each UIS is sometimes accurate, but sometimes quite inaccurate. The factors underlying those differences are described and discussed.

The rules for MYC contain implicit assumptions; a fact which Shortliffe and Buchanan very briefly discuss at two places in [Shortliffe 76]. The *and* and *or* rules for MYC operate on confidence factors which denote not belief but *changes* in belief. Thus, we need to do some algebra in order to state what the assumptions are, but the end result is that the *and* and *or* operations are equivalent to Jeffrey's Rule, but using only one piece of the given data. Suppose that $cf(A_1) \leq 0$ and $cf(A_1) \leq cf(A_2)$, then equation 9 holds. Given the definition of a MYC CF as a re-scaling, we can restate 9 as 10, and follow the algebra to the last line, which as one can see is exactly the answer which MXE gives if we use only $p_1(A_1)$ and ignore evidence about $A_2$.

$$cf(A_1 \text{ and } A_2) = cf(A_1) \tag{9}$$

$$\tag{10}$$

A similar derivation gives equation 11, when $0 \leq cf(A_1) \leq cf(A_2)$. For this equation, the odds-ratio answer is identical only under the additional assumption that $A_1$ is a proper sub-set of $A_2$ (or, equivalently, that $A_2$ logically implies $A_1$. Also, equation 11 is exactly the MXE result if $p(A_2)$ is 1.0, implying that MYC will often over-estimate the probability of a conjunction when both CF's increase. Note however, that this extra assumption is invoked only some of the time, and depends not on prior probabilities but on how much they are changed and in which direction.

$$p_1(\neg(A_1 \& A_2)) = p_0(\neg(A_1 \& A_2)) \frac{p_1(\neg A_1)}{p_0(\neg A_1)} \tag{11}$$

For the *or* operation, we may also do similar algebra. For the case that $cf(A_1) \geq 0$ and $cf(A_1) \geq cf(A2)$, we get equation 12, which is exactly the MXE answer, if we ignore evidence about $A_2$. For the case that $0 \geq cf(A_1) \geq cf(A_2)$, we get equation 13, which is the MXE answer if we ignore evidence about $A_2$ and assume that $A_2$ is a proper subset $A_1$. Again, this additional assumption is only invoked sometimes.

$$p_1(\neg(A_1 \, or \, A_2)) = p_0(\neg(A_1 \, or \, A_2)) \frac{p_1(\neg A_1)}{p_0(\neg A_1)} \tag{12}$$

$$p_1(A_1 \, or \, A_2) = p_0(A_1 \, or \, A_2) \frac{p_1(A_1)}{p_0(A_1)} \tag{13}$$

The fact that MYC updates with only one piece of data while systematicly ignoring the rest has certain implications. For example, one would expect MYC to under-estimate the impact of new data. In fact, if one takes the data presented in [Shortliffe 76], comparing Mycin CF's to the correct probabilities, and does a linear regression, one finds that MYC's response is only about 51% of the correct amount (explaining 74% of the variance). When there are N pieces of input data, MYC's *and* and *or* operators will ignore N-1 of them. This can cause extreme errors. For example, suppose $p_0(A_1)$ is quite near 1.0, and $p_0(A_i)$ is near 0.0 for $2 \leq i \leq N$; the prior probability of their conjunction will generally be nearly zero. If $cf(A_1)=0$, and $cf(A_i)=1$ for all other i, then MYC will estimate the cf of their conjunction as 0.0, and hence estimate the posterior probability of the conjunction to remain at nearly zero, even though it has actually risen to be quite near 1.0. In fact, it is exactly $p_0(A_1)$. Also, in the face of contradictory data, MYC will not balance one against the other, but simply update with one or the other. Updating with one will have an impact in the right direction; updating with the other will have an impact in the wrong direction.

We compared the *and* rule for MYC and TSM, on two input items, to the results of *and* in minimum cross entropy updates. Surprisingly, we found that the following results were true whether the propositions were positively correlated, negatively correlated, or independent: if both input CF's are negative, MYC understates the impact of the data, and if either one or both of the input CF's are positive, MYC overstates the impact of the data. The analysis was done as follows. We assumed a prior distribution in which the prior probabilities of $A_1$ and $A_2$ are both one half, and the prior probability of $A_1$ & $A_2$ is one ninth. The results for this particular distribution are given in table 3-1; results for other distributions with negative correlation were qualitatively similar.

| $CF(A_1)$ | $CF(A_2)$ | $CF_{myc}(A_1 \& A_2)$ | $CF_{mxe}(A_1 \& A_2)$ |
|---|---|---|---|
| +.8 | +.8 | +.8 | +.776 |
| -.8 | +.8 | -.8 | -.501 |
| -.8 | -.8 | -.8 | -.991 |

**Figure 3-1:** Negative Correlation Comparison of MYC and MXE "*and*" rule

If we set their CF's to +0.8 and +0.8, this gives posterior probabilities of 0.9 to $A_1$ and 0.9 to $A_2$. If we do an MXE update with those new probabilities, we find a posterior probability for $(A_1 \& A_2)$ of 0.801, which given the prior of one ninth, corresponds to a CF of +0.776 for the conjunction. MYC's value of +0.8 CF for the conjunction corresponds to a posterior probability of 0.822. In this case, MYC has overstated the impact by only 3.08%, which may be negligible. If we set their CF's to +0.8 and -0.8, this gives posterior probabilities of 0.9 to $A_1$ and 0.1 to $A_2$. If we do an MXE update with those new probabilities, we find a posterior probability for $(A_1 \& A_2)$ of 0.05543, corresponding to a CF of -.501. The -0.8 CF for the conjunction corresponds to a posterior probability of 0.022. In this case, MYC has overstated the impact by 59.6%, which would probably be significant. If we set their CF's to -0.8 and -0.8, this gives posterior probabilities of 0.1 to $A_1$ and 0.1 to $A_2$. If we do an MXE update with those new probabilities, we find a posterior probability for $(A_1 \& A_2)$ of 0.001, or a CF of -.991. The -0.8 CF for the conjunction corresponds to a posterior probability of 0.022. In this case, MYC has understated the impact.

These results are simply explained if one bears in mind that MYC updates using Jeffrey's Rule, but ignores all save one piece of data. In the case of CF's with opposite signs, the impact of the data is over-stated because only the disconfirming evidence is used. The confirming evidence, which should have lessened the impact, is totally ignored. When both have negative signs, the impact is under-stated, because MYC ignores the fact that it has multiple pieces of disconfirming evidence, and acts as if it only had one. When both are positive, MYC also assumes that one, say $A_1$, is a logical consequence of the other, $A_2$, as shown by equation 11. If that "assumed rule" were true, then $A_2$ itself would actually be additional evidence for $A_1$. Of course, this is not true in our case, and so MYC overstates the impact, because it thinks it has received more information than it really has. Equivalently, MYC treats $A_2$ as having probability 1.0, and hence it overstates the probability of $(A_1 \& A_2)$.

This line of reasoning implies that the higher the positive correlation between $A_1$ and $A_2$, the less significant will be the underestimation when both CF's are negative. Similarly, we expect the over-estimation to be greater when both are positive and there is positive correlation, because then the two CF's are more redundant, and hence less informative. Both hypotheses are confirmed by the experiments on a distribution in which $A_1$ and $A_2$ are positively correlated (i.e. $p(A_1)=p(A_2)=.5$, $p(A_1 \& A_2)=.389$), with the results in figure 3-2.

| $CF(A_1)$ | $CF(A_2)$ | $CF_{myc}(A_1 \& A_2)$ | $CF_{mxe}(A_1 \& A_2)$ |
|---|---|---|---|
| +.8 | +.8 | +.8 | +.746 |
| +.8 | -.8 | -.8 | -.746 |
| -.8 | -.8 | -.8 | -.885 |

**Figure 3-2:** Positive Correlation Comparison of MYC and MXE "*and*" rule

325

We may similarly compare MYC's *or* rule to the MXE results. Again, and for similar reasons MYC understates the impact of the data whenever both CF's are positive, but the underestimation decreases when $A_1$ and $A_2$ increase in correlation. For *or*, MYC always overestimates the impact of data whenever there are CF's of mixed sign, because the disconfirming evidence is ignored. Whenever both CF's are negative, it makes the same assumption as before that one antecedent is a logical consequence of the other, and hence behaves as if there were an additional rule, which exaggerates the impact.

The MYC rule for combining the results of multiple rules which bear on one consequent is rather complex. Because it is quadratic in the input CF's, which are non-linearly related to the probabilities, inverting and solving in the fashion of equations 10 through 13 yields equations which have no clear intuitive interpretation. Hence, we choose numerical analysis. We compared it to the ordinary *or* rule of MYC and the *or* results of MXE, for the previous two prior distributions over $A_1$ and $A_2$, as well as one in which they were independent. In table 3-3, the prior distribution is flat over all possible combinations of $A_1$ and $A_2$, which makes them independent with probability 0.5 each. The table displays the CF's for $A_1$ and $A_2$ at the far left. The CF's yielded by using the MYC *or* and *rule-or* rule are in the middle. On the far right is the CF obtained for ($A_1$ or $A_2$) if one converts the two input CF's to posterior probabilities, does the MXE update, reads the probability of the disjunction off the posterior, and converts that back to a CF (using the prior probability of the disjunction).

| $CF(A_1)$ | $CF(A_2)$ | $CF_{myc-or}$ | $CF_{rule-or}$ | $CF_{mxe-or}$ |
|---|---|---|---|---|
| +.8 | +.8 | +.8 | +.96 | +.9600 |
| +.8 | -.8 | +.8 | 0.00 | +.6400 |
| +.4 | +.4 | +.4 | +.64 | +.6400 |
| -.8 | -.8 | -.8 | -.96 | -.7467 |

**Figure 3-3:** Independence Comparison of MYC "or", MYC "rule-or", and MXE "or" rules

Whenever the CF's are of equal magnitude but opposite signs, the *or* result overestimates the MXE result, both of them are positive, and the *rule-or* result predicts no change at all. Whenever both CF's are positive the *rule-or* result is exactly the MXE result, and *or* underestimates the impact of the data. But when both are negative, the *rule-or* and *or* results both overestimate the impact of the data, which is a puzzling asymmetry. We will now argue that the *rule-or* rule is equivalent to assuming independence when both CF's are positive, and that there is a strong negative correlation when both CF's are negative.

Let us suppose that $CF(A) \geq 0$ and $CF(B) \geq 0$. Recalling the definitions of positive CF's and of *rule-or*, we get equation 14. For brevity, we have designated $p_0(A)$ by $A_0$, $p_1(A)$ by $A_1$, and so on. This equation states that the CF for (A or B), assuming that they are independent, is the *rule-or* combination of their individual CF's. Surprisingly, the second line of equation 14 is an identity, true for any choice of $A_0$, $A_1$, $B_0$, and $B_1$ with $A_0 \neq 1$ and $B_0 \neq 1$. Hence, it is also true for any choice which satisfies $0 \leq A_0 \leq A_1 \leq 1.0$ and $0 \leq B_0 \leq B_1 \leq 1.0$. One should note that the assumption of independence was something which MYC was explicitly designed to avoid. Instead, it seems merely to have been obscured by MYC's notation, illustrating the point that without careful analysis, it is not clear what a heuristic UIS really does, or even whether it does what it was designed to do.

$$cf(A \text{ rule-or } B) = cf(A) + cf(B) - cf(A)cf(B) \tag{14}$$

$$\frac{(A_1+B_1-A_1B_1)-(A_0+B_0-A_0B_0)}{1-(A_0+B_0-A_0B_0)} = \frac{A_1-A_0}{1-A_0} + \frac{B_1-B_0}{1-B_0} - \frac{(A_1-A_0)(B_1-B_0)}{(1-A_0)(1-B_0)}$$

If we take the same approach when both CF's are negative, we get equation 15. But the second line can not be satisfied if $A_1 \leq A_0$ and $B_1 \leq B_0$, because the left side of the equality will always be larger than the right.

$$cf(A \text{ rule-or } B) = cf(A) + cf(B) + cf(A)cf(B) \tag{15}$$

$$\frac{1}{A_1} + \frac{1}{B_1} = \frac{1}{A_0} + \frac{1}{B_0}$$

Hence, for negative CF's, *rule-or* does not model independence. In fact, if one combines the *or* results from the distributtions with negative, zero, and positive correlation which we mentioned earlier, one can see from table 3-4 that *rule-or* corresponds to assuming a very strong negative correlation. This is particularly significant because the commonly accepted assumption of conditional independence always produces non-negative correlation.

| CF(A) | CF(B) | rule-or | $MXE_{minus}$ | $MXE_{indep}$ | $MXE_{plus}$ |
|-------|-------|---------|---------------|---------------|--------------|
| -.8   | -.8   | -.96    | -.776         | -.747         | -.746        |

**Figure 3-4:** Comparing rule-or to various degrees of correlation

Most authors require that an inference system obey DeMorgan's Laws, in equation 16. At first, it seems that MYC's and, *or*, and *not* rules obey them. However, the CF's refer to *change* in belief, not to belief itself. Hence, it is not clear from their definitions alone whether MYC will actually obey DeMorgan's Laws in terms of belief. But the evidence above clearly shows that the *rule-or* and *or* operations, *when combined* with the non-linear definition of CF's, generally violate DeMorgan's Laws. This is because it assumes independence when CF(A) and CF(B) are both positive, but if we reverse the signs of the CF's by talking about CF($\neg$A) and CF($\neg$B), then it assumes a strong negative correlation. Hence, the results in the first case will directly contradict those from the second.

$$A \text{ or } B = \neg(\neg A \text{ \& } \neg B) \qquad A \text{ \& } B = \neg(\neg A \text{ or } \neg B) \tag{16}$$

Clearly, we have not really changed the content of the data-base, just re-named the propositions within it. Thus, because the operations are sensitive to the *signs* of the CF's, MYC will make completely different implicit assumptions, and get very different answers, purely because the propositions have been re-named. One will recall that invariance under 1-to-1 one transformations was one of the desirable, defining characteristics of MXE updating. Not only does the MYC result depend on trivial restatements of the prior, but also it makes different, contradictory, assumptions about the prior depending on which update is performed.

The results of the various detailed comparisons may be integrated and summarized as follows. Shortliffe and Buchanan's original MYC data shows an overall tendency to under-estimate the impact of new data by about 49%. Our more detailed analysis shows that the *and, or*, and hence also *modus ponens* rules systematically ignore data and hence typically underestimate the aggregate impact. Only the operation for combining the results of several rules consistently produces over-estimations of the impact, and it may have a significant bias toward positive CF's. Hence, we may conjecture that in the complete rule-system which Shortliffe and Buchanen tested, the under-estimation effect

327

of the first three operations dominated the over-estimation effect of the last one. This suggests that efforts to improve the MYC system are better directed at the way it does *and*, *or*, and *modus ponens* than at the way it combines the results of separate rules. Even though the latter issue may have more intellectual appeal than the first three, it seems to have less practical importance.

### 4. Experiments on Performance

While we have clarified some biases in MYC and TSM, we have not yet addressed the problem of how important those biases really are. To answer this question, we examined their $\zeta$-performance over a broad range of differently structured sample rule-sets, and multiple trials of each rule-set. Space limitations preclude complete discussion of the data, but the general method is more important. It is important to note that our tests do cover a broad range of basic structures. This type of broad coverage and comparison of results has been lacking in the few previous studies which examined UIS's performance.

For each rule-set, a collection of input data were evaluated, and each UIS assigned its average performance - the average value of $\zeta$ - over that set of cases. The rule-sets varied between one and twelve rules. The average was found by by cycling each input node through a range of four probabilities. Thus, if there were $n$ input leaves, there would be $4^n$ cases tried, one for each possible combination of the inputs. The four values were always close to 0.05, 0.35, 0.65, 0.95. They were actually chosen to be random within .01 of those values; for example, (0.053,0.349,0.643,0.945) would be a possible set of values. They were generated afresh for each of the $4^n$ trials. This sampling method was chosen for two reasons. First, it spanned the space of input possibilities, and thus approximated an even distribution over the space of possible input probabilities. Second, it enabled trees with different numbers of input nodes to have their performance compared, as only the averages were compared. Thus, the numbers derived are an estimate of average performance, which is distinct from the biases, which were explored earlier. The small examples came in eight sets, totaling 36 small examples. Each example was run for between 16 and 256 individual trials, for each UIS (MYC, TSM, Prospector, Fuzzy Sets, CI, Independence Assumptions, and PULS). We will present some data on MYC's and TSM's performance over the first six sets. In each set, one or several factors were changed in systematic way to see the effect of different structures.

Depth: The first two examples (dpth-2 and dpth-1) tested effects of depth. This was done by setting up a tree two rules deep and solving for the ME solution, giving the prior for dpth-2. The resulting probabilities for the intermediate nodes, $B_1$ and $B_2$, were then included as constraints on a rule-set which was just the top rule of the first set, giving the prior for dpth-1. Separate series of experiments were then run on each, giving average $\zeta$ performance over the top node, A.

Rule Strength and Bushiness: The second four examples were each just one rule. They tested of varying bushiness and of varying lower rule strength. The first case, bsh2-upr, just has the upper rule strength for A given the conjunction of $B_1$ and $B_2$. We add a lower rule strength in bsh2-u&l, a third antecedent, $B_3$, in case bsh3-upr, and both the lower rule strength and third antecedent in case bsh3-u&l. Thus, the four cases form a two by two matrix of options.

Shared antecedents: The third set of six examples tested some simple correlations. Cases 2cnc-2rls-neg and 2cnc-2rls-pos had two consequents, $A_1$ and $A_2$, each referred to by its own rule. $A_1$ depended on $B_1$ and $B_2$, while $A_2$ depended on $B_2$ and $B_3$, so the rules shared an antecedent. In case 2cnc-2rls-pos, this antecedent was shared directly, creating positive correlation between $A_1$ and $A_2$. In 2cnc-2rls-neg, the antecedent has its negation used in the second rule, creating a negative correlation between $A_1$ and $A_2$.
In cases 1cnc-2rls-neg and 1cnc-2rls-pos, use the same two rules to make conclusions about one shared consequent, A. Similarly to the first two cases, the two rules in 1cnc-2rls-pos share the same antecedent, and so the satisfaction of their compound antecedent terms is positively correlated, while in case 1cnc-2rls-neg, the negation is shared, and so the satisfaction is negatively correlated. Cases 1cnc-2lyrs-neg and 1cnc-2lyrs-pos are just like 2cnc-2rls-neg and 2cnc-2rls-pos, respectively, except that another rule has been added, so that the final conclusion is separated by one more layer of rules from the source of correlation. One should note that 1cnc-2rls-neg, 1cnc-2rls-pos, 1cnc-2lyrs-neg, and 1cnc-2lyrs-pos all contain undirected cycles, and hence cannot be perfectly modeled by the CI system's assumptions.

<u>Shared Conclusions</u>: The fourth set of two examples (cnd-ind-2 and cnd-ind-3) test the effect of varying bushiness in cases where several rules bear on one consequent. As derived in an appendix to [Wise 86], in the ME prior conditional independence holds exactly for such cases.

<u>Rule Strength, Bushiness, and Correlated Inputs</u>: The fifth set of examples (bsh2-upr-pos bsh2-upr-neg bsh2-u&l-pos bsh2-u&l-neg bsh3-upr-pos bsh3-upr-neg bsh3-u&l-pos bsh3-u&l-neg) test the effects of varying bushiness, lower rule strength, and correlation of inputs. They reproduce the two-by-two experimental design of the second set of examples, but add the extra factor of explicitly introducing either the minimum or maximum possible correlation of the antecedents, giving a two-by-two-by-two design.

<u>Extreme Correlations of Inputs</u>: The sixth set of four examples test the effects of varying correlation of consequents and of rules. In cases 2cnc-min-shr-ruls-pos 2cnc-max-shr-ruls-pos, the rules share an antecedent, and so instances of their being satisfied are positively correlated; in cases 2cnc-min-shr-ruls-neg 2cnc-max-shr-ruls-neg, negatively. In cases 2cnc-min-shr-ruls-pos and 2cnc-min-shr-ruls-neg, the consequents have the minimum possible overlap; in cases 2cnc-max-shr-ruls-pos and 2cnc-max-shr-ruls-neg, the maximum possible. Hence, we again have a two-by-two design.

## 5. Best and Worst Results for MYC and TSM

The three best and three worst $\zeta$-performances for MYC are listed in figure 5-1; 1.0 is the best possible; -1.0 is the worst. From figure 5-2., we can see that sensitivity to rule strength should outweigh the sensitivity to antecedent correlations. The higher the upper rule strength, the better MYC should do. The performance on cases where the lower rule strength was explicitly forced down (e.g. 5.8 and 5.7) confirms this. Similarly, bsh2-upr-neg and bsh3-upr-neg have no lower rule strength specified, which lets them drift higher, and MYC does quite well.

| $\zeta$ | Case name | Case Number |
|---|---|---|
| .998 | 1cnc-2lyrs-neg | 3.5 |
| .996 | bsh2-upr-neg | 5.1 |
| .994 | bsh3-upr-neg | 5.2 |
| .788 | bsh3-u&l-pos | 5.8 |
| .735 | bsh2-u&l-pos | 5.7 |
| .517 | cnd-ind-2 | 4.1 |

**Figure 5-1:** Best and Worst Results for MYC

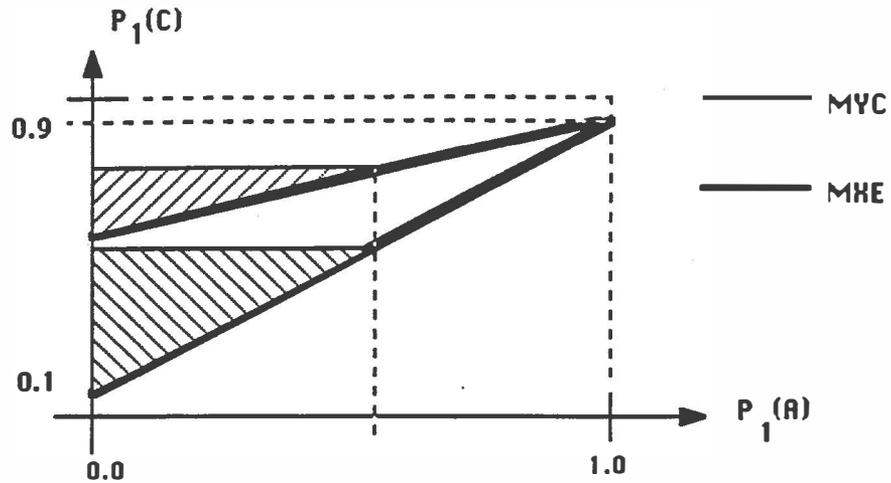
**Figure 5-2:** Sensitivities for MYC vs. MXE

The TSM system does read the rule-strengths off the prior, hence the only effects on accuracy are those which pertain to the correlations of inputs. Note that most of the best cases were with only two antecedents rather than with three (as is typical of the worstcases). Recalling that TSM ignores all but one of the input items, this accords well with our expectation that it would do worse when ignoring 2/3 of the data than when only ignoring 1/2 of it.

| $\zeta$ | Case name | Case Number |
|---|---|---|
| .997 | cnd-ind-2 | 4.1 |
| .983 | 2cnc-min-shr-ruls-neg | 6.2 |
| .981 | 2cnc-min-shr-ruls-pos | 6.1 |
| | | |
| .138 | bsh3-upr | 2.3 |
| .128 | bsh3-upr-neg | 5.2 |
| .011 | bsh3-upr-pos | 5.6 |

**Figure 5-3:** Best and Worst Results for TSM

Another way of looking at these results is suggested by the fact that TSM's worst four cases were all among MYC's best cases. Moreover, TSM's average performances on its four best and four worst cases are lower than the corresponding averages for MYC. This is because MYC essentially has "canceling errors". The problem appears when we get a mixture of confirming and disconfirming evidence. Because it responds to only one piece of data and ignores cancellation, TSM tends to over-react in one direction or the other when mixed data is present. However, MYC stays put at the prior value. In many cases, it is better to not react at all than to over-react in the wrong direction. This is an example of how fixing only one error (no sensitivity to disconfirming data), but not the other (using only one piece of data in Jeffrey's Rule) can actually worsen performance.

Table 5-4 lists the worst $\zeta$-performance for each of the UIS's discussed here, and gives some information about robustness. Explaining the differing degrees of robustness provides a useful, extremely brief, summary of the relevant interactions and biases for each UIS.

| CI | MYC | TSM |
|---|---|---|
| +.643 | +.517 | +.011 |

**Figure 5-4:** Worst performance for each UIS

One can see that the UIS with the best worst-$\zeta$ was CI, suggesting that it is the most robust of those tested here. This is simply because CI uses more parameters than do the other UIS's, and hence has more degrees of freedom to "bend to fit the data" even in cases where its assumptions are unmet. For example, it can model antecedants' as having negative, zero, or positive correlation. In spite of having only one operation different, TSM performed worse than did MYC, because it only corrects one of two canceling errors. Of the six UIS's tested, these three were the best; the others all had average performance which was worse than random guessing (i.e. $\zeta \leq 0$) on at least some cases.

## 6. Summary

Our theoretical predictions of bias are born out by numerical analysis of performance. Thus, we have gained some insight into what the relative strengths of the effects are. Also, we have seen how some deleterious, or meritorious, conditions arise for subtle reasons, hence we have more clues to check for while trying to estimate how well a UIS will do in a given application. For Mycin and its variant, we found special situations where its performance was very good, but also situations where performance was worse than random guessing, or where data was interpreted as having the opposite of its true import. We have uncovered independence assumptions in MYC, inspite of the fact that eliminating such assumptions was one of the designers' goals. We have found that all three of these systems usually gave accurate results, and that the conditional independence assumptions gave the most robust results. Considerations of robustness might be a critical factor in selecting UIS's for a given application.

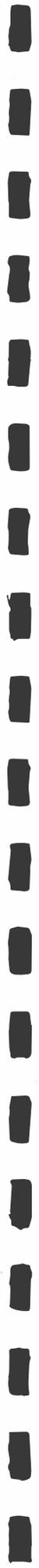